\documentclass[11pt,letterpaper]{article}
\usepackage{emnlp2016}
\usepackage{times}
\usepackage{latexsym}
\usepackage{tabularx}
\usepackage{url}
\usepackage{graphicx}
\usepackage{amssymb,amsmath}
\usepackage{bm}
\usepackage{makecell}
\usepackage{multi row}
\usepackage[font=footnotesize,labelfont=bf]{caption}

\emnlpfinalcopy 
\def\emnlppaperid{1008}

\usepackage{etoolbox}
\newcommand{\zerodisplayskips}{
  \setlength{\abovedisplayskip}{8pt}
  \setlength{\belowdisplayskip}{8pt}
  \setlength{\abovedisplayshortskip}{4pt}
  \setlength{\belowdisplayshortskip}{4pt}}
\appto{\normalsize}{\zerodisplayskips}
\appto{\small}{\zerodisplayskips}
\appto{\footnotesize}{\zerodisplayskips}

\newcommand*{\Resize}[2]{\resizebox{#1}{!}{$#2$}}
\newcommand{\alns}[1] {
	\begin{align*} #1 \end{align*}
}

\newcommand{\bh} { \bm{h} }
\newcommand{\bW} { \bm{W} }
\newcommand{\bw} { \bm{w} }
\newcommand{\bb} { \bm{b} }
\newcommand{\mT} { \mathcal{T} }
\newcommand{\mC} { \mathcal{C} }
\newcommand{\mA} { \mathcal{A} }
\newcommand{\delr} { \Delta_{r} }
\newcommand{\loss} { \mathcal{L(\theta)} }
\newcommand{\reward} { J(\theta) } 
\newcommand{\pt} { p_\theta }
\DeclareMathOperator*{\argmax}{argmax}
\DeclareMathOperator{\E}{\mathbb{E}}
\DeclareMathOperator{\dt}{\nabla_{\theta}}
\newcommand{\overbar}[1]{\mkern 1mu\overline{\mkern-1mu#1\mkern-9mu}\mkern 9mu}

\newcommand\Tstrut{\rule{0pt}{2.6ex}}
\newcommand\Bstrut{\rule[-1.0ex]{0pt}{0pt}}
\newcommand{\mline}{\Xhline{1.0\arrayrulewidth}}
\newcommand{\tline}{\Xhline{2.5\arrayrulewidth}}
\newcommand{\lone}	{\Tstrut \Bstrut \\ \mline}
\newcommand{\ttop}{\tline}
\newcommand{\tbottom}{\Bstrut \\ \tline}

\newcommand{\xhdr}[1]{\vspace{1.7mm}\noindent{{\bf #1.}}}

\title{Deep Reinforcement Learning for Mention-Ranking Coreference Models}
\author{
	Kevin Clark\\
	Computer Science Department\\
	Stanford University \\
	{\tt kevclark@cs.stanford.edu}
\And
	Christopher D. Manning \\
  	Computer Science Department \\
  	Stanford University \\
  	{\tt manning@cs.stanford.edu} \\}
\date{}

\begin{document}

\maketitle

\begin{abstract}
Coreference resolution systems are typically trained with heuristic loss functions that require careful tuning. In this paper we instead apply reinforcement learning to directly optimize a neural mention-ranking model for coreference evaluation metrics. We experiment with two approaches: the {\sc reinforce} policy gradient algorithm and a reward-rescaled max-margin objective. We find the latter to be more effective, resulting in significant improvements over the current state-of-the-art on the English and Chinese portions of the CoNLL 2012 Shared Task.
\end{abstract}

\section{Introduction}
Coreference resolution systems typically operate by making sequences of local decisions (e.g., adding a coreference link between two mentions). However, most measures of coreference resolution performance do not decompose over local decisions, which means the utility of a particular decision is not known until all other decisions have been made.

Due to this difficulty, coreference systems are usually trained with loss functions that heuristically define the goodness of a particular coreference decision. These losses contain hyperparameters that are carefully selected to ensure the model performs well according to coreference evaluation metrics. This complicates training, especially across different languages and datasets where systems may work best with different settings of the hyperparameters. 

To address this, we explore using two variants of reinforcement learning to directly optimize a coreference system for coreference evaluation metrics. In particular, we modify the max-margin coreference objective proposed by Wiseman et al. (2015) by incorporating the reward associated with each coreference decision into the loss's slack rescaling. We also test the {\sc reinforce} policy gradient algorithm \cite{williams1992simple}.

Our model is a neural mention-ranking model. Mention-ranking models score pairs of mentions for their likelihood of coreference rather than comparing partial coreference clusters. Hence they operate in a simple setting where coreference decisions are made independently. Although they are less expressive than entity-centric approaches to coreference (e.g., Haghighi and Klein, 2010\nocite{haghighi2010coreference}), mention-ranking models are fast, scalable, and simple to train, causing them to be the dominant approach to coreference in recent years \cite{durrett2013easy,wiseman2015learning}. Having independent actions is particularly useful when applying reinforcement learning because it means a particular action's effect on the final reward can be computed efficiently. 

We evaluate the models on the English and Chinese portions of the CoNLL 2012 Shared Task. The {\sc reinforce} algorithm is competitive with a heuristic loss function while the reward-rescaled objective significantly outperforms both\footnote{Code and trained models are available at \url{https://github.com/clarkkev/deep-coref}.}.
We attribute this to reward rescaling being well suited for a ranking task due to its max-margin loss as well as benefiting from directly optimizing for coreference metrics. Error analysis shows that using the reward-rescaling loss results in a similar number of mistakes as the heuristic loss, but the mistakes tend to be less severe.

\section{Neural Mention-Ranking Model} \vspace{-0mm}
We use the neural mention-ranking model described in Clark and Manning (2016)\nocite{clark2016improving}, which we briefly go over in this section. Given a mention $m$ and candidate antecedent $c$, the mention-ranking model produces a score for the pair $s(c, m)$ indicating their compatibility for coreference with a feedforward neural network. The candidate antecedent may be any mention that occurs before $m$ in the document or {\sc na}, indicating that $m$ has no antecedent. \\

\xhdr{Input Layer}
For each mention, the model extracts various words (e.g., the mention's head word) and groups of words (e.g., all words in the mention's sentence) that are fed into the neural network. Each word is represented by a vector $\bw_i\in \mathbb{R}^{d_w}$. Each group of words is represented by the average of the vectors of each word in the group. In addition to the embeddings, a small number of additional features are used, including distance, string matching, and speaker identification features. See Clark and Manning (2016) \nocite{clark2016improving} for the full set of features and an ablation study.  

These features are concatenated to produce an $I$-dimensional vector $\bh_0$, the input to the neural network. If $c = \text{\sc na}$, features defined over pairs of mentions are not included. For this case, we train a separate network with an identical architecture to the pair network except for the input layer to produce anaphoricity scores. 

\xhdr{Hidden Layers}
The input gets passed through three hidden layers of rectified linear (ReLU) units \cite{nair2010rectified}. Each unit in a hidden layer is fully connected to the previous layer:
\[
	\bh_i(c, m) = \max(0, \bW_i \bh_{i -1}(c, m) + \bb_i)
\]
where $\bW_1$ is a $M_1 \times I$ weight matrix, $\bW_2$ is a $M_2 \times M_{1}$ matrix, and $\bW_3$ is a $M_3 \times M_2$ matrix. 

\xhdr{Scoring Layer}
The final layer is a fully connected layer of size 1:
\[
	s(c, m) = \bW_{4} \bh_3(c, m) + b_4
\]
where $\bW_4$ is a $1 \times M_3$ weight matrix. At test time, the mention-ranking model links each mention with its highest scoring candidate antecedent.

\section{Learning Algorithms}
Mention-ranking models are typically trained with heuristic loss functions that are tuned via hyperparameters. These hyperparameters are usually given as costs for different error types, which are used to bias the coreference system towards making more or fewer coreference links. 
 
In this section we first describe a heuristic loss function incorporating this idea from Wiseman et al. (2015)\nocite{wiseman2015learning}.
We then propose new training procedures based on reinforcement learning that instead directly optimize for coreference evaluation metrics.

\subsection{Heuristic Max-Margin Objective}
The heuristic loss from Wiseman et al. is governed by the following error types, which were first proposed by Durrett et al. (2013)\nocite{durrett2013decentralized}.

 Suppose the training set consists of $N$ mentions $m_1, m_2, ..., m_N$. Let $\mC(m_i)$ denote the set of candidate antecedents of a mention $m_i$ (i.e., mentions preceding $m_i$ and \textsc{na}) and $\mathcal{T}(m_i)$ denote the set of true antecedents of $m_i$ (i.e., mentions preceding $m_i$ that are coreferent with it or $\{\textsc{na}\}$ if $m_i$ has no antecedent). Then we define the following costs for linking $m_i$ to a candidate antecedent $c \in \mC(m_i)$:
\[
    \Delta_h(c, m_i)= 
\begin{cases}
    \alpha_{\textsc{fn}}& \text{if }c = \textsc{na } \wedge \mT(m_i) \neq \{\textsc{na}\}\\
    \alpha_{\textsc{fa}}& \text{if }c \neq \textsc{na } \wedge \mT(m_i) = \{\textsc{na}\}\\
    \alpha_{\textsc{wl}}& \text{if }c \neq \textsc{na } \wedge c \notin \mT(m_i) \\
    0 & \text{if } c \in \mT(m_i)
\end{cases}
\]
for ``false new,'' ``false anaphor,'' ``wrong link'', and correct coreference decisions.

The heuristic loss is a slack-rescaled max-margin objective parameterized by these error costs. Let  $\hat{t}_i$ be the highest scoring true antecedent of $m_i$:
\[
	\hat{t}_i = \underset{c \in \mathcal{C}(m_i) \wedge \Delta_h(c, m_i) = 0}\argmax s(c, m_i)
\]

\noindent Then the heuristic loss is given as
\[\Resize{0.5\textwidth}{
	\loss = \sum\limits_{i=1}^N \underset{c \in \mC(m_i)}\max  \Delta_h(c, m_i)(1 + s(c, m_i) - s(\hat{t}_i, m_i))
}\]

\xhdr{Finding Effective Error Penalties} We fix $\alpha_{\textsc{wl}} = 1.0$ and search for $\alpha_{\textsc{fa}}$ and $\alpha_{\textsc{fn}}$ out of $\{0.1, 0.2, ..., 1.5\}$ with a variant of grid search.
Each new trial uses the unexplored set of hyperparameters that has the closest Manhattan distance to the best setting found so far on the dev set. The search is halted when all immediate neighbors (within 0.1 distance) of the best setting have been explored. 
We found  $(\alpha_{\textsc{fn}}, \alpha_{\textsc{fa}}, \alpha_{\textsc{wl}}) = (0.8, 0.4, 1.0)$ to be best for English and  $(\alpha_{\textsc{fn}}, \alpha_{\textsc{fa}}, \alpha_{\textsc{wl}}) = (0.8, 0.5, 1.0)$ to be best for Chinese on the CoNLL 2012 data.

\subsection{Reinforcement Learning}
Finding the best hyperparameter settings for the heuristic loss requires training many variants of the model, and at best results in an objective that is correlated with coreference evaluation metrics. To address this, we pose mention ranking in the reinforcement learning framework \cite{sutton1998reinforcement} and propose methods that directly optimize the model for coreference metrics. 

We can view the mention-ranking model as an {\it agent} taking a series of {\it actions} $a_{1:T} = a_1, a_2, ..., a_T$, where $T$ is the number of mentions in the current document. Each action $a_i$ links the $i$th mention in the document $m_i$ to a candidate antecedent. 
Formally, we denote the set of actions available for the $i$th mention as $\mA_i = \{(c, m_i): c \in \mC(m_i)\}$, where an action $(c, m)$ adds a coreference link between mentions $c$ and $m$. The mention-ranking model assigns each action the score $s(c, m)$ and takes the highest-scoring action at each step.

 Once the agent has executed a sequence of actions, it observes a {\it reward} $R(a_{1:T})$, which can be any function. We use the B$^3$ coreference metric for this reward \cite{bagga1998algorithms}. Although our system evaluation also includes the MUC \cite{vilain1995model} and CEAF{$_{\phi_4}$ \cite{luo2005coreference} metrics, we do not incorporate them into the loss because MUC has the flaw of treating all errors equally and CEAF{$_{\phi_4}$ is slow to compute. 

\xhdr{Reward Rescaling}
Crucially, the actions taken by a mention-ranking model are independent. This means it is possible to change any action $a_i$ to a different one $a_i' \in \mA_i$ and see what reward the model would have gotten by taking that action instead: $R(a_1,...,a_{i - 1},a_i',a_{i + 1},...,a_T)$. We use this idea to improve the slack-rescaling parameter $\Delta$ in the max-margin loss $\loss$. Instead of setting its value based on the error type, we compute exactly how much each action hurts the final reward:
\alns{
\delr(c, m_i) = -&R(a_1,...,(c, m_i),...,a_T) \\
	+ &\max \limits_{a_i' \in \mA_i} R(a_1,...,a_i',...,a_T)
}
where $a_{1:T}$ is the highest scoring sequence of actions according to the model's current parameters. Otherwise the model is trained in the same way as with the heuristic loss. 

\xhdr{The REINFORCE Algorithm}
We also explore using the {\sc reinforce} policy gradient algorithm \cite{williams1992simple}. We can define a probability distribution over actions using the mention-ranking model's scoring function as follows:
\[
	\pt(a) \propto e^{s(c, m)}
\]
for any action $a = (c, m)$. 
The {\sc reinforce} algorithm seeks to maximize the expected reward
\alns{
	\reward &=  \E_{[a_{1:T} \sim \pt]}R(a_{1:T})
}
It does this through gradient ascent. Computing the full gradient is prohibitive because of the expectation over all possible action sequences, which is exponential in the length of the sequence. Instead, it gets an unbiased estimate of the gradient by sampling a sequence of actions $a_{1:T}$ according to $\pt$ and computing the gradient only over the sample. 

We take advantage of the independence of actions by using the following gradient estimate, which has lower variance than the standard {\sc reinforce} gradient estimate:
\[\Resize{0.5\textwidth}{
	\dt \reward \approx \sum\limits_{i=1}^T \sum\limits_{a_i' \in \mA_i}[\dt \pt(a_i')](R(a_1,...,a_i',...,a_T) - b_i)
}\]
where $b_i$ is a baseline used to reduce the variance, which we set to $\E\limits_{a_i' \in \mA_i \sim \pt}R(a_1,...,a_i',...,a_T)$.

\begin{table*}[tb]
 \vspace{-0mm}
\small
\setlength{\tabcolsep}{4pt}
\centering
\begin{tabular}{l@{\hskip 10pt} c c c@{\hskip 17pt}   c c c@{\hskip 17pt}   c c c@{\hskip 11pt}   c} 
\ttop
 & \multicolumn{3}{c}{MUC} &  \multicolumn{3}{c}{B$^3$} & \multicolumn{3}{c}{CEAF$_{\phi_4}$} & \Tstrut \\ 
 & Prec. & Rec. & $F_1$ & Prec. & Rec. & $F_1$ & Prec. & Rec. & $F_1$ & Avg. $F_1$ \Bstrut\\ \tline
\multicolumn{11}{c}{\textbf{CoNLL 2012 English Test Data}}  \Tstrut\Bstrut\\ \tline
 Wiseman et al. (2016) & 77.49 & 69.75 & 73.42 & 66.83 & 56.95 & 61.50 & 62.14 & 53.85 & 57.70 & 64.21\Tstrut \\
 Clark \& Manning (2016)  & 79.91 & 69.30 & 74.23 & 71.01 & 56.53 & 62.95 & 63.84 & 54.33 & 58.70 & 65.29\Bstrut \\  \mline
 Heuristic Loss   & 79.63 & 70.25 & \textbf{74.65} & 69.21 & 57.87 & 63.03 & 63.62 & 53.97 & 58.40 & 65.36\Tstrut\\
 {\sc reinforce}    & 80.08 & 69.61 & 74.48 & 70.70 & 56.96 & 63.09 & 63.59 & 54.46 & 58.67 & 65.41\\
 Reward Rescaling  & 79.19 & 70.44 & 74.56 & 69.93 & 57.99 &  \textbf{63.40} & 63.46 & 55.52 &  \textbf{59.23} & \textbf{65.73}\tbottom
 \multicolumn{11}{c}{\textbf{CoNLL 2012 Chinese Test Data}}  \Tstrut\Bstrut\\ \tline
 Bj{\"o}rkelund \& Kuhn  (2014)  & 69.39 & 62.57 & 65.80 & 61.64 & 53.87 & 57.49 & 59.33 & 54.65 & 56.89 & 60.06 \Tstrut \\
 Clark \& Manning (2016)  & 73.85 & 65.42 & 69.38 & 67.53 & 56.41 & 61.47 & 62.84 & 57.62 & 60.12 & 63.66 \Bstrut \\  \mline
 Heuristic Loss   & 72.20 & 66.51 & 69.24 & 64.71 & 58.16 & 61.26 & 61.98 & 58.41 & 60.14 & 63.54\Tstrut\\
 {\sc reinforce}    & 74.05 & 65.38 & \textbf{69.44} & 67.52 & 56.43 & 61.48 & 62.38 & 57.77 & 59.98 & 63.64 \\
 Reward Rescaling    & 73.64 & 65.62 & 69.40 & 67.48 & 56.94 & \textbf{61.76} & 62.46 & 58.60 & \textbf{60.47} & \textbf{63.88}\tbottom
 \end{tabular}
\caption{\small Comparison 
of the methods together with other state-of-the-art approaches on the test sets. }
 \vspace{-3mm}
\label{tab:final}
\end{table*} 
\nocite{bjorkelund2014learning,wiseman2016learning}

\section{Experiments and Results}
We run experiments on the English and Chinese portions of the CoNLL 2012 Shared Task data \cite{pradhan2012conll} and evaluate with the MUC, B$^{3}$, and CEAF{$_{\phi_4}$ metrics. 
Our experiments were run using predicted mentions from Stanford's rule-based coreference system \cite{raghunathan2010multi}. 

We follow the training methodology from Clark and Manning (2016): hidden layers of sizes $M_1$ = 1000, $M_2$ = $M_3$ = 500, the RMSprop optimizer \cite{rmsprop}, dropout \cite{hinton2012improving} with a rate of 0.5, and pretraining with the all pairs classification and top pairs classification tasks. However, we improve on the previous system by using using better mention detection, more effective hyperparameters, and more epochs of training.

\subsection{Results}
We compare the heuristic loss, {\sc reinforce}, and reward rescaling approaches on both datasets. We find that {\sc reinforce} does slightly better than the heuristic loss, but reward rescaling performs significantly better than both on both languages. 

We attribute the modest improvement of {\sc REINFORCE} to it being poorly suited for a ranking task. During training it optimizes the model's performance in expectation, but at test-time it takes the most probable sequence of actions. This mismatch occurs even at the level of an individual decision: the model only links the current mention to a single antecedent, but is trained to assign high probability to all correct antecedents. We believe the benefit of {\sc REINFORCE} being guided by coreference evaluation metrics is offset by this disadvantage, which does not occur in the max-margin approaches. The reward-rescaled max-margin loss combines the best of both worlds, resulting in superior performance.

\begin{figure}[tb]
\begin{center}
\vspace{-2mm}
\includegraphics[width=0.49\textwidth]{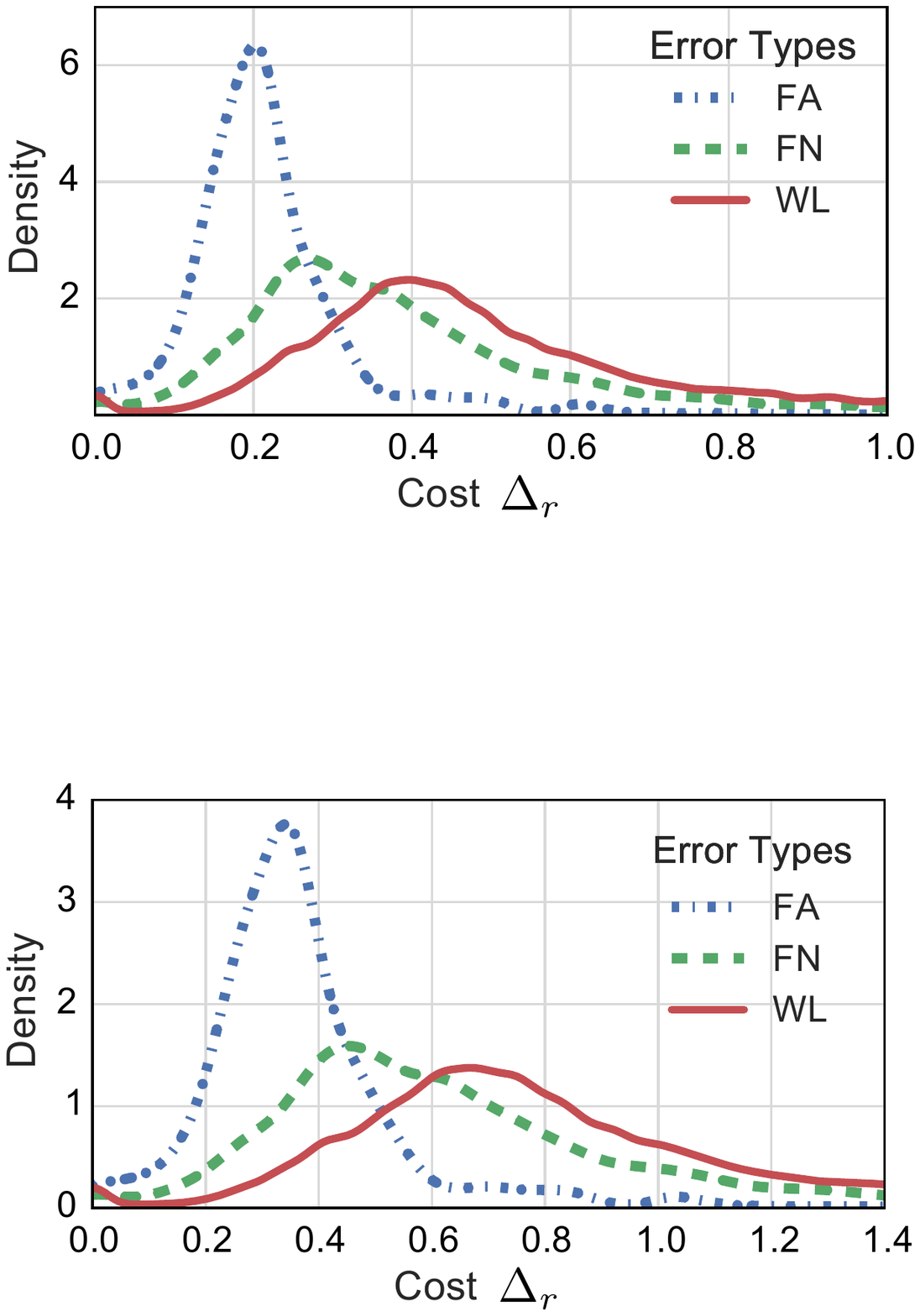}
\end{center}
\vspace{-2mm}
\caption{Density plot of the costs $\delr$ associated with different error types on the English CoNLL 2012 test set.}
\vspace{-3mm}
\label{fig:error_costs}
\end{figure}

\subsection{The Benefits of Reinforcement Learning}

In this section we examine the reward-based cost function $\delr$ and perform error analysis to determine how reward rescaling improves the mention-ranking model's accuracy. 

\xhdr{Comparison with Heuristic Costs}
We compare the reward-based cost function $\delr$ with the error types used in the heuristic loss. For English, the average value of $\delr$ is 0.79 for \textsc{fn} errors and 0.38 for \textsc{fa} errors when the costs are scaled so the average value of a \textsc{wl} error is 1.0. These are very close to the hyperparameter values $(\alpha_{\textsc{fn}}, \alpha_{\textsc{fa}}, \alpha_{\textsc{wl}}) = (0.8, 0.4, 1.0)$ found by grid search. However, there is a high variance in costs for each error type, suggesting that using a fixed penalty for each type as in the heuristic loss is insufficient (see Figure 1). 

\begin{table}[tb]
\begin{tabularx}{\columnwidth}{>{\centering\arraybackslash}p{3cm} >{\centering\arraybackslash}p{1cm} >{\centering\arraybackslash}p{1cm} >{\centering\arraybackslash}p{1cm} }
\ttop 
Model & \textsc{fn}  & \textsc{fa} & \textsc{wl} \lone
Heuristic Loss & 1719 & 1956 & 1258\Tstrut\\
Reward Rescaling & 1725 & 1994 & 1247\tbottom
\end{tabularx}
\caption{Number of ``false new," ``false anaphoric," and ``wrong link" errors produced by the models on the English CoNLL 2012 test set.}
\vspace{-4mm}
\label{tab:errors}
\end{table} 

\begin{table*}[tb]
\small
\setlength{\tabcolsep}{-1pt}
\centering
\tabcolsep=3pt
\begin{tabular}{l@{\hskip 7mm} c c c@{\hskip 18mm}   c c c@{\hskip 18mm}   c c >{\centering\arraybackslash}p{6mm}} 
\ttop
 \multirow{2}{*}{Mention Type} & \multicolumn{3}{l}{\hspace{-1mm}Average Cost $\overbar{\delr}$} &  \multicolumn{3}{l}{\hspace{-8mm}\# Heuristic Loss Errors} & \multicolumn{3}{l}{\hspace{-8mm}\# Reward Rescaling Errors} \Tstrut \\ 
 & \sc{fn} & \sc{fa} & \sc{wl} & \sc{fn} & \sc{fa} & \sc{wl} & \sc{fn} & \sc{fa} & \sc{wl} \Bstrut\lone
Proper nouns & 0.90 & 0.38 & 1.02 & 403 & 597 & 221 & 334 & 660 & 233\Tstrut\\
Pronouns in phone conversations & 0.86 & 0.39 & 1.21 & 82 & 85 & 81 & 90 & 78 & 67\tbottom
 \end{tabular}
\caption{Examples of classes of mention on which the reward-rescaling loss improves upon the heuristic loss due to its reward-based cost function. Reported numbers are from the English CoNLL 2012 test set. } 
 \vspace{-2mm}
\label{tab:costs}
\end{table*} 

\xhdr{Avoiding Costly Mistakes}
Embedding the costs of actions into the loss function causes the reward-rescaling model to prioritize getting the more important coreference decisions (i.e., the ones with the biggest impact on the final score) correct. As a result, it makes fewer costly mistakes at test time. Costly mistakes often involve large clusters of mentions: incorrectly combining two coreference clusters of size ten is much worse than incorrectly combining two clusters of size one. However, the cost of an action also depends on other factors such as the number of errors already present in the clusters and the utilities of the other available actions.

Table~\ref{tab:errors} shows the breakdown of errors made by the heuristic and reward-rescaling models on the test set. The reward-rescaling model makes slightly more errors, meaning its improvement in performance must come from its errors being less severe. 

\xhdr{Example Improvements}
Table~\ref{tab:costs} shows two classes of mentions where the reward-rescaling loss particularly improves over the heuristic loss. 

Proper nouns have a higher average cost for ``false new'' errors (0.90) than other mentions types (0.77). This is perhaps because proper nouns are important for connecting clusters of mentions far apart in a document, so incorrectly linking a proper noun to \textsc{na} could result in a large decrease in recall. Because it more heavily weights these high-cost errors during training, the reward-rescaling model makes fewer ``false new'' errors for proper nouns than the heuristic loss. Although there is an increase in other kinds of errors as a result, most of these are low-cost ``false anaphoric" errors.

The pronouns in the ``telephone conversation" genre often group into extremely large coreference clusters, which means a ``wrong link'' error can have a large negative effect on the score. This is reflected in its high average cost of 1.21. After prioritizing these examples during training, the reward-rescaling model creates significantly fewer wrong links than the heuristic loss, which is trained using a fixed cost of 1.0 for all wrong links. 

\section{Related Work}
Mention-ranking models have been widely used for coreference resolution \cite{denis2007ranking,rahman2009supervised,durrett2013easy}. 
These models are typically trained with heuristic loss functions that assign costs to different error types, as in the heuristic loss we describe in Section 3.1 \cite{fernandes2012latent,durrett2013decentralized,bjorkelund2014learning,wiseman2015learning,martschat2015latent,wiseman2016learning}. 

To the best of our knowledge reinforcement learning has not been applied to coreference resolution before. However, imitation learning algorithms such as SEARN \cite{daume2009search} have been used to train coreference resolvers \cite{daume2006practical,ma2014prune,clark2015entity}. These algorithms also directly optimize for coreference evaluation metrics, but they require an expert policy to learn from instead of relying on rewards alone. 

\section{Conclusion}
We propose using reinforcement learning to directly optimize mention-ranking models for coreference evaluation metrics, obviating the need for hyperparameters that must be carefully selected for each particular language, dataset, and evaluation metric. Our reward-rescaling approach also increases the model's accuracy, resulting in significant gains over the current state-of-the-art.

\section*{Acknowledgments}
We thank Kelvin Guu, William Hamilton, Will Monroe, and the anonymous reviewers for their
thoughtful comments and suggestions. This work was supported by NSF Award
IIS-1514268.

\bibliography{clark-manning-emnlp16-deep}
\bibliographystyle{emnlp2016}

\end{document}